# Visual Self-supervised Learning Scheme for Dense Prediction Tasks on X-ray Images

Shervin Halat
Computer Engineering Department
Amirkabir University of Technology
Tehran, Iran
shervin.halat@aut.ac.ir

Mohammad Rahmati
Computer Engineering Department
Amirkabir University of Technology
Tehran, Iran
rahmati@aut.ac.ir

Ehsan Nazerfard
Computer Engineering Department
Amirkabir University of Technology
Tehran, Iran
nazerfard@aut.ac.ir

*Abstract*—Recently, significant advancements in artificial intelligence have been attributed to the integration of self-supervised learning (SSL) scheme. While SSL has shown impressive achievements in natural language processing (NLP), its progress in computer vision has comparatively lagged behind. However, the incorporation of contrastive learning into existing visual SSL models has led to considerable progress, often surpassing supervised counterparts. Nonetheless, these improvements have been mostly limited to classification tasks. Moreover, few studies have evaluated visual SSL models in real-world scenarios, as most have focused on datasets with class-wise portrait images, notably ImageNet. Here, we focus on dense prediction tasks using security inspection x-ray images to evaluate our proposed model, Segment Localization (SegLoc). Based upon the Instance Localization (InsLoc) model, SegLoc addresses one of the key challenges of contrastive learning, i.e., false negative pairs of query embeddings. Our pre-training dataset is synthesized by cutting, transforming, and pasting labeled segments from an existing labeled dataset (PIDray) as foregrounds onto instances from an unlabeled dataset (SIXray) as backgrounds. Furthermore, we fully leverage the labeled data by incorporating the concept – one queue per class – into the MoCo-v2 memory bank, thereby avoiding false negative pairs. In our experiments, SegLoc outperformed random initialization by 3% to 6% while underperformed supervised initialization, in terms of AR and AP metrics across different IoU values over 20 to 30 pre-training epochs.

## I. Introduction

Astonishing advancements achieved in Natural Language Processing (NLP) by incorporation of Self-supervised Learning (SSL) scheme have sparked innovations in SSL methods in computer vision as well. Indeed, large language models such as BERT [1], GPT [2], and many others have gained so much popularity, which is culminated with the introduction of OpenAI's ChatGPT in the last few months, leading to a paradigm shift in AI models. In spite of acknowledged achievements of SSL in NLP, there are still complications regarding SSL in computer vision. Early introduced visual SSL methods were simply basic variations on existing ideas implemented in that of natural language models whereby a part of or the whole input, being text or image, was masked or transformed, and the model in return, was expected to restore the undistorted input or determine the applied transformation given the distorted version of the input. Models introduced in [3-8] can be mentioned in this regard.

Inconsistency between SSL performance in NLP and computer vision may be attributed to dimensional discrepancies between the two areas [9]. Not being able to fulfill expectations, these early visual SSL models have further been modified by being implemented in contrastive learning manner, basically in Siamese fashion, in view of the fact that high-level encoded representations of various transformations of identical view should remain almost the same. This idea was incompatible with previous practices; as in non-contrastive methods, including [4], the models were expected to infer the angular rotation or applied transformation on the corresponding initial images.

Superior performance of contrastive learning based visual SSL methods has clearly illustrated credibility of aforementioned arguments. Despite all achievements however, there are still unsolved challenges in contrastive models which are yet to be addressed, most notable ones being:

- underperformance in dense prediction tasks compared to classification task, which may be justifiable considering the intrinsic nature of contrastive learning.

- false negative pairs; as in contrastive learning, regardless of context, instances are mutually treated as negative pairs to each other, thereby dedicating a distinct class per instance by disregarding intra- and inter class correlations. This argument can be further justified considering the superiority of cluster-based SSL methods, namely [11] and [12], where each outperformed its own counterparts. This may be inferred from table 5 of paper [13] and figure 3 of paper [14] for classic and contrastive based SSL models, respectively.

- lack of a comprehensive study on efficacy of different SSL methods in specialized tasks and datasets, medical, satellite, security inspection, and microscopic images to name but a few. Hence, cross-domain generalizability of existing methods is still a matter of debate.

To address the first aforementioned challenge, model Instance Localization (InsLoc) [14] has managed to incorporate localization task into contrastive learning method, model MoCo-v2 [15] in this case, through novelty of cropping a pair of random patches of a randomly selected image from unlabeled dataset in question, then randomly transforming each patch to generate two different views (foreground images), which in turn are pasted on another randomly selected image acting as background image. This procedure, in its basic form, exploits carefully curated nature of ImageNet dataset [16], which is implicitly biased toward datasets containing portrait images with only single object covering the majority of image; however, this is not the case in most of wild, natural, or specialized images.

Taking all aforementioned challenges into account, we have come up with a seamless extension of Instance Localization [14], namely Segment Localization, Fig. 1. The main contributions of our method along with its corresponding arguments are as follows:

*1) Segment based copying and pasting for dataset preparation of pre-training:* The core contribution of our method to address claimed issues linked to InsLoc model is to prepare foregrounds from segments of labeled dataset of downstream task in question. Indeed, without prior knowledge, the simple act of crop and paste has some drawbacks from different perspectives which could potentially offset its benefits if not override it in many real-world tasks. First and foremost, the rationale behind viability of such practice becomes invalidated where there are dense and crowded images as this practice may result in obscure and meaningless synthesized pair of images, or even worse, supposedly positive views containing objects of different classes. Regarding our chosen dataset, a positive pair synthesized by the approach suggested in Instance Localization [14] is illustrated in Fig. 3 along with that of our approach Segment Localization in Fig. 2 depicting two different positive pairs. Furthermore, almost all objects have contour-like shapes, as against boxes, which may further deteriorate the functionality of InsLoc model; therefore, cropping even a random contour may be even more beneficiary than boxes; additionally, the inconsistency between copied samples and the background along the edges of the patch may lead the model to seek for trivial solutions owing to the box-shaped artifacts. Consequently, here we propose substituting random box-shaped patches with real segments extracted from labeled images from already available segmented datasets (as foregrounds), or considering downstream labeled dataset - in our case, PIDray dataset [17], which is a security inspection x-ray image dataset with objects of interest being prohibited items. In addition, regarding backgrounds, any dataset can be considered with similar domain and context as that of downstream task's dataset - in our case, SIXray dataset [18] is considered, the vast majority of images of which being benign (i.e., not containing downstream objects of interest, prohibited items in our case). Our approach offers more control over the learning process with other associated benefits being addressing some principal challenges of contrastive learning, that is, applying the contrastive learning on a pair of images each of which containing distinct object sample from an identical object class (i.e., two samples of different brands or models from equal class); in addition, being aware of each segment's class label, we can be genuinely assured of avoiding false negative samples, as against previous contrastive learning based models, which is addressed through the following initiative.

*2) Dictionary maintenance adaption of MoCo model [9] (One Queue per Class):* Taking labeled segments as foregrounds, we have managed to confidently discard any possible false negative pairs corresponding to query

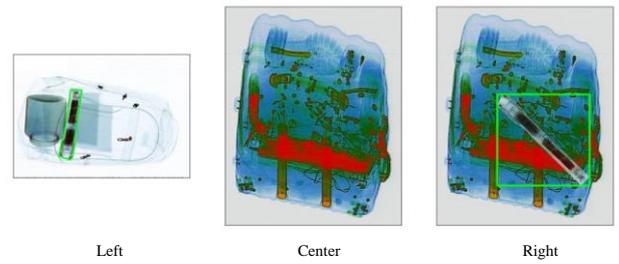

Left  Center  Right

Fig. 1. Segment Localization – contrastive learning-based SSL model by cutting segments of already labeled instances (foreground) on unlabeled instances (backgrounds). Left: already labeled instance with baton segment in x-ray image. Center: readily available unlabeled benign x-ray image. Right: transformed segment pasted on background instance.

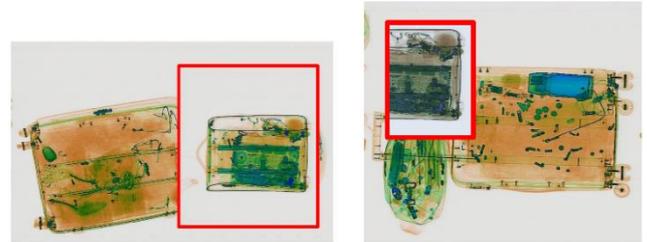

Fig. 2. A pair of positive samples synthesized with InsLoc method on SIXray dataset. (Foregrounds, shown in red bounding boxes, are transformed from the same instance and pasted on different backgrounds.)

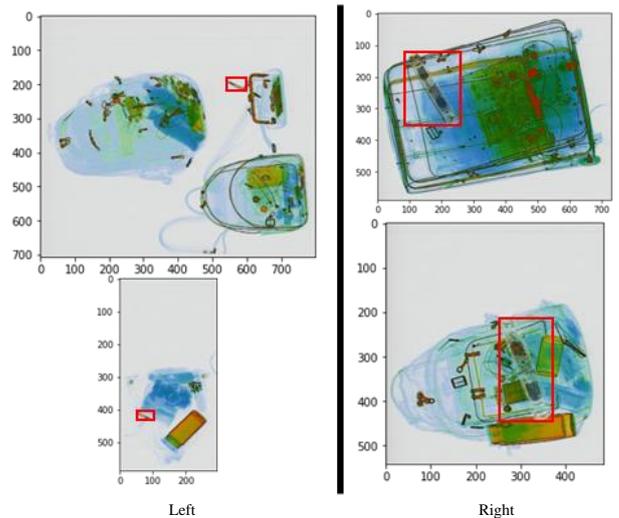

Left  Right

Fig. 3. Two samples of positive pairs synthesized by our approach SegLoc –Left: a positive pair containing bullet. Right: a positive pair containing baton.

representation at hand through maintaining distinct queue per class. To do so, we maintain independent queues each linked to a separate class so that each queue only contains keys corresponding to samples of classes other than that of queue's class. The schematic model of this adaption is shown in Fig. 4.

*3) Evaluation of the proposed scheme on a real-world scenario (i.e., Security Inspection of x-ray images):* To evaluate our model, we have selected dense prediction task of semantic segmentation of prohibited items in security inspection x-ray images. This selection is rooted in the fact that, first and foremost, this domain differs from that of natural images which has been the case in most prior models. Secondly, this task can be viewed as a classic example of domain-specific real-world scenario with specialized datasets where massive amounts of unlabeled data are being generated daily without getting harnessed due to expensive labeling process, both qualitatively and quantitatively, being a suitable target for self-supervised learning scheme. Lastly, dense prediction tasks of security inspection x-ray images are exceptionally challenging both in terms of technical aspects, with objects being densely packed along with transparency present in x-ray images where objects' outlines overlap in addition to elimination of appearance and texture of objects, and security risks associating with prohibited items.

*4) Discarding grayscaling and bounding box augmentations while adding rotation transformation:* We argue that applicability of existing SSL models in real-world scenarios is not straightforward and requires non-trivial refinements beyond architectural alignments [14]. For instance, color is of crucial importance in security inspection x-ray images as different objects appear in distinct colors depending on materials and substances being made of. Additionally, angular orientation plays a critical role in such a task as representations are demanded to be invariant to orientations while being equivariant at the same time regarding detection and segmentation tasks, respectively. Moreover, by virtue of pasting genuine segments, the proposed bounding box augmentation of InsLoc model is no longer needed in our model Segment Localization (SegLoc) as backgrounds are naturally present at margins of segments' bounding boxes, and in turn, two advantages accrued from this elimination, being elimination of, first, exhaustive search for overlapping anchor boxes, and second, any possible detrimental effect of bounding box augmentation, as bounding boxes are expected to be bounded to objects' limits, which is overlooked in InsLoc in this regard. Accordingly, we have discarded grayscaling and bounding box augmentations while adding rotation transformation into our model in contrast to InsLoc.

*5) Incorporation of SDANet as the backbone of our model:* Being able to fairly evaluate and compare our model's performance, and also, SDANet being the SOTA model for its corresponding task as of the time being published [17], we have utilized SDANet for our model.

Aside from technicalities, we hope that this work and ideas implemented function as a stepping stone for new researchers interested in entering this field of research but are being deterred by resource-intensive aspects of self-supervised learning.

## II. BACKGROUND

The core idea of self-supervised learning is to exploit supposedly useless massive amounts of unlabeled data being prepared on a daily basis in different modalities and domains with the aim of reinforcing models to extract more meaningful high-level features. Harnessing unlabeled data roots back in much earlier under settings being called weakly supervised, semi-supervised and unsupervised learning. Despite significant overlaps, outstanding advancements and considerations in recent years required self-supervised learning paradigm to have its distinct name.

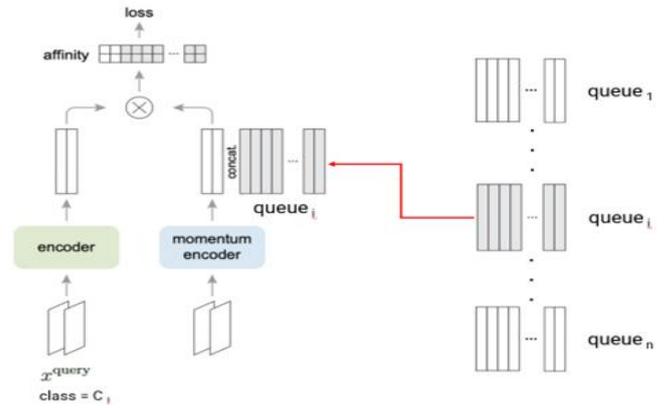

Fig. 4. One Queue per Class - Our adaption to dictionary maintenance proposed by MoCo method. (The artwork is extended from Fig. 1 of MoCo-v2 paper [15])

There are some notable comprehensive surveys on visual self-supervised learning, including the paper [12] which provides an in-depth introduction to early and classic self-supervised learning methods as of 2019. Also, a recent paper titled Cookbook of SSL [19] has provided an invaluable resource to lower the entrance barrier for new researchers in SSL, specifically contrastive based visual SSL. In this paper, they bridge the classic approaches to more recent contrastive and/or vision transformer [20] based approaches along with trying to democratizing self-supervised learning scheme. In classic approaches models were generally trained to estimate the applied augmentation or restore missing or distorted part of unlabeled input image under unsupervised fashion by considering the original image as pseudo-label.

Despite lowering the gap between unsupervised and supervised learning performance, still, classic methods underperformed by their supervised counterparts. To boost viability of these methods, contrastive self-supervised learning methods, such as [11, 21-23], were introduced with their main objective being enforcing invariance of representations corresponding to different transformations or views of the same instance in contrastive learning manner. A survey on contrastive learning gives an introduction to different architectures and loss functions of earlier contrastive learning models employed in SSL [13] as well as [19]. These contrastive models can be categorized mainly based on type of loss functions employed and how positive and/or negative pair(s) of corresponding instance is/are obtained. Hence, these models typically can be categorized as Deep Metric Learning family such as SimCLR [22], Self-Distillation family such as BYOL [23], MoCo [1-9, 15], and Dino [24], and Canonical Correlation Analysis family such as VICReg [25] and SwAV [11].

Underperformance of earlier contrastive based SSL models in visual dense prediction tasks encouraged more innovative research focused on such tasks. Accordingly, some models have just been proposed on top of existing contrastive baselines to enhance their performance in tasks of detection and/or segmentation for the most part, including models, such as DetCo [26], UP-DETR [27], SoCo [28], and InsLoc [14], to name but a few. More precisely, the most popular choice in terms of contrastive baseline has been MoCo which comes in three versions v1 [9], v2[15], v3 [29] with v1 and v2 difference being in addition of projection head and transformations applied, also, v3 has discarded memory queue while picking keys from concurrent batch, resembling SimCLR, along with adding prediction head on top of query encoder.

Nevertheless, comparatively, very few number of studies have been conducted on real-world scenarios, notably in domains other than natural images. However, studies such as [30], [31], and [32] can be pointed out in this regard. More broadly, this paper [33] has intensively surveyed on works done in the intersection of SSL and medical imaging analysis.

Concerning ideas employed in our work Segment Localization, in [31] they have proposed model MICLe in which they effectively improve accuracy of medical image classification by two consecutive SSL pre-training, with first one being on ImageNet and the next one on domain-specific medical image dataset while harnessing metadata of patients in order to provide real positive pairs. More pricesly, instead of augmenting each instance to generate positive pairs, they have managed to provide corresponding positive pair by taking another image taken from the same pathology of the very same patient based on metadata. In essence, this has meaningful overlaps with our work both in terms of multi-instance approach for contrastive learning and the fact that they have pre-trained MICLe then fine-tuned their model by the same dataset which has been the case in our approach. Additionally, in [32], they have successfully managed to extract more meaningful discriminative generic person appearance descriptor by contrasting that of each person to its corresponding genuine positive and negative pairs achieved by exploitation of three conditions, i.e., motion tracking, mutual exclusion constraints, and multi-view geometry. Finally, in InsLoc [14], which has served as our model's baseline, has leveraged the idea of RoIAlign [34] in order to address prior models' architectural misalignment between upstream and downstream models for dense prediction tasks, and also incorporate image patches into contrastive learning. More pricesly, this has achieved by pasting two randomly cut patches of a randomly selected image onto another randomly selected image, images which served as foreground and background, respectively, providing synthesized positive pairs.

Regarding our test case, prohibited items in security inspection x-ray images, the work [18], as of its release in early 2019, introduced the largest security inspection x-ray images SIXray at the time.

Their dataset contains more than 1 million x-ray images, of size 100,000 pixels on average, of real baggages and luggages; however, only a fraction of a percent (0.84%) of the dataset includes prohibited items, 8,929 out of 1,059,231 instances. Consequently, eliminating labeled part of the dataset, the rest has served us as our unlabeled benign background images not containing any kind of prohibited items.

In addition, in late 2021, [17] introduced PIDray dataset of the same domain and specialty as SIXray, which has served as largest of high quality labeled dataset in task of prohibited items detection in security inspection x-ray images considering diversity, quality, and quantity altogether; More precisely, PIDray dataset contains 47,677 images containing prohibited items of 12 distinct classes (with 2 to 15 unique instances of each class), where each image's prohibited item is finely located and segmented. In this work, we have employed PIDray dataset to extract our intended foreground segments as well as both fine-tuning and testing.

## III. PRETEXT TASK – SEGMENT LOCALIZATION

Our proposed scheme Segment Localization requires task-specific data preparation settings prior to self-supervised pre-training. Taking our specific task into consideration/, we have set a few measures, thereby synthesizing more legitimate inputs, including authentic region determination and composition coefficient.

### A. Pre-training Data Pre-processing

In our approach, in order to perform the contrastive learning, training image pairs are synthesized such that each pair consists of two images containing distinct objects of the same class. That is, both images within each pair has common class label.

Here the process of data preparation for each sample pair is discussed. First, two random images are selected from the background dataset (SIXray dataset in our case), then, authentic region of each are determined. Although, this practice can be performed in advance by determining authentic region per background image, which has been the case for our experiments for the sake of reducing training time. Then, randomly, one of the possible object classes contained in foreground dataset (PIDray in our case) will be selected through which two random images containing object of the same class are randomly selected from PIDray dataset as well. Next, intended segments of each image are cut and transformed being able to be pasted on the background. In our specific case, the pasting is done by randomly specifying a value within the range of (0.25, 0.65) named as composition coefficient. The choice of composition coefficient value range is on view of the fact that objects in x-ray images tend to overlap depending on their corresponding absorption and reflection rate along with their positioning relative to each other. The actual precise coefficient range to be used is beyond the scope of this research; hence, the proposed range is roughly estimated through visually analyzing synthesized images and qualitatively sifting. Lastly, transformed segments are pasted onto a random point within corresponding background's authentic region. The pseudo-code of the pre-training dataset synthesis is given in Algorithm 1.

### B. Authentic Region Determination

As mentioned, for each background image, an authentic region is estimated so as to ensure validation of synthesized images, that is to avoid segments being pasted on unreasonable whitish margins of x-ray images. Indeed, most of SIXray x-ray images' regions are empty areas which are appeared in noisy whitish regions in images. Hence, by determining the authentic region, we get assured that the transformed segments are to be pasted on a valid non-empty region (indicating bags, luggage, or any other type of container). To automatize the process, only a rough estimation without much computational overhead has been considered sufficient. Accordingly, beginning from four sides, as if the whole image is authentic at the beginning of the process, by continuously shrinking the region from all four sides simultaneously, each side of the region will be brought to standstill once reaching the first pixel of an object. Consequently, a box-shaped region is obtained demonstrating an approximation of the authentic region, shown in Fig. 5. The explained process has been implemented in the following manner which involves multiple consecutive image processing steps:

*1) RGB to Grayscale Conversion:* The image is converted from RGB to Grayscale.

*2) Erosion:* The obtained grayscale image is eroded by a kernel of size 9. In fact, whitish background of x-ray images is subject to scattered weak noises which tend to significantly deteriorate determination of authentic region unless being eliminated.

*3) Thresholding:* The grayscale format of the image is thresholded (by value of 50 in our specific case, which has been obtained by trial and error) to limits of 0 and 255 indicating non-object and object pixel, respectively.

*4) Authentic Region:* Eventually, the bounding box containing non-zero pixels will be considered as corresponding authentic region.

## IV. LEARNING APPROACH

Aside from data synthesis procedure during pre-training which does not affect the training process by itself, our learning approach is basically similar to that of InsLoc method regarding integration of bounding boxes along with RoIAlign module into contrastive learning framework.

In terms of conceptual and architectural modifications, there has been a nuance concerning the maintenance of dictionary queue which keeps track of negative keys pairs for queries. Also, in order for our approach to be kept aligned with the downstream task's architecture, the backbone introduced in [17], namely SDANet, is integrated into our model. The mentioned modifications are further discussed in the following:

### A. Dictionary Maintenace Adaption

In order to maximize potential of availability of segments' class labels, we have proposed the idea of maintaining an exclusive dictionary queue per class whereby, at the end of each training step, each queue will be updated only by keys corresponding to images of current batch not containing objects of its class. Consequently, representations of images containing the same objects will be implicitly attracted to each other in the feature space by being explicitly contrasted with true negative keys within the corresponding queue. The schematic architecture of the proposed dictionary is depicted in Fig. 4.

### B. Backbone Alignment

In [17], in order to modify the segmentation of Cascade Mask R-CNN model [35] for the specialized task of prohibited item detection in x-ray images, they have proposed SDANet having an attention module integrated into its ResNet-101 backbone; Therefore, in order for us to seamlessly transfer pre-trained model to downstream task, we have employed SDANet backbone as the backbone of our own contrastive learning model. Fig. 6 shows final architecture of our pre-training model.

## V. EXPERIMENTS

The evaluation process of our proposed scheme is based on conventional settings considered in other self-supervised methods through which the performance of the proposed model is compared with that of random initialization along

**Algorithm 1. Pseudocode of pre-training data synthesis**

```
# backs = background images dataset
# fores = foreground images dataset
# fores_labels = foreground images labels
# authentic_region = function to find authentic region

# randomly taking two background images
back_1, back_2 = random(backs, 2)

# determination of background images authentic region
region_1 = authentic_region(back_1)
region_2 = authentic_region(back_2)

# randomly selecting one of the classes
class_num = random(number_of_classes)

# randomly taking two foreground images
fore_1, fore_2 = random(fores, 2)

# extracting intended segments
seg_1, seg_2 = cut(fore_1, fore_2, fores_labels)

# transforming segments
trans_seg_1, trans_seg_2 = transform(seg_1, seg_2)

# pasting transformed segments on background authentic regions
query, label_1 = paste(trans_seg1, back_1, region_1)
positive_key, label_2 = paste(trans_seg2, back_2, region_2)

# bundling two synthesized images to create a positive pair
positive_pair = (query, label_1, positive_key, label_2)
```

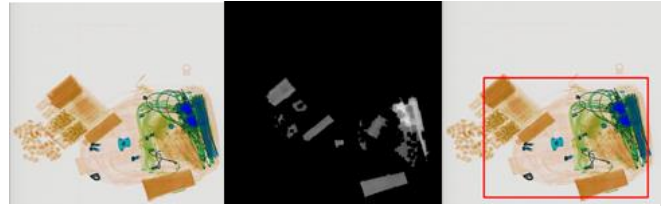

Fig. 5. Authentic region determination sample – Left: initial background image. Center: after transformation with grayscaling, erosion, and thresholding. Right: determined authentic region for segments to be pasted on.

with supervised initialization. Specifically, each of the two comparisons is evaluated on object detection and semantic segmentation on test set of PIDray dataset, where it consists of three categories Easy, Hard, and Hidden, covering 40% of the dataset combined. These subsets indicate different difficulty levels of detection, that is, Easy subset includes images containing only one object to be recognized, Hard subset includes images containing at least two objects to be recognized, and Hidden subset includes more realistic images that the prohibited items were deliberately hidden, thereby being more realistic and closer to real-world scenarios. Specifically, in our task, after conducting the SegLoc pre-training, the pre-trained backbone is transferred to the downstream model (SDANet in our case), then, the SDANet model will be fine-tuned in the exact same manner as that of its paper [17]. In the following, the configured settings implemented in each training phase are pointed out. Then, the obtained results are given, and lastly, the outcomes of ablation studies are discussed. It should be noted that demonstrated results for SDANet are regenerated by us. Also, the metrics employed in the experiments are AP, mAP, AR, and mAR based on COCO standard.

The pre-training is performed by synthesized dataset of size 200,000 labeled with segments' class and bounding box. The synthesized dataset is obtained through Algorithm 1. with unlabeled portion of SIXray dataset along with training set of PIDray dataset (60% of the whole dataset). Synthesized images are resized to 500 approximately with regard to width size while retaining aspect ratio; therefore, the input images are not of the same size. The contrastive learning is implemented in the same manner as MoCo except for its dictionary that there are one queue corresponding to each segment class with each queue being of size 16385 (i.e., $2^{14}$). Also, experiments are conducted with mini-batches of size 64.

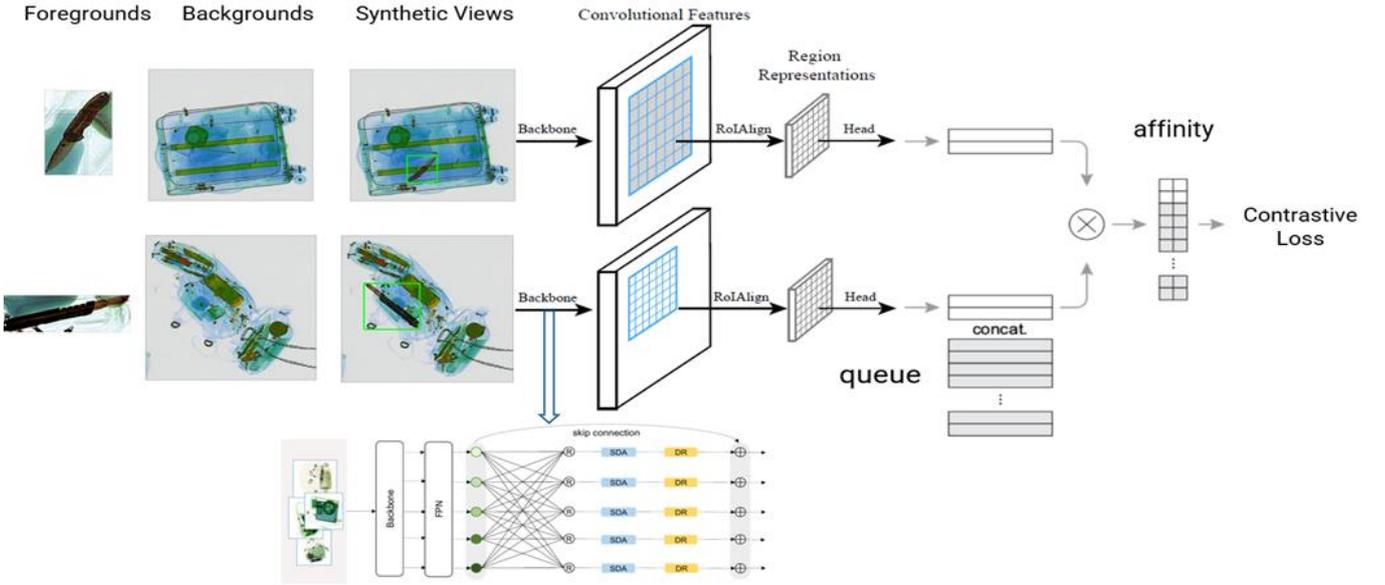

Fig. 6. Final architecture of our model SegLoc (the artwork is extended from Fig. 3 of [15] and Fig. 5 of [2])

Overall, pre-training has been performed for about 30 epochs with the optimizer being the same as InsLoc settings. For pre-training process, a 12 gigabyte GeForce GTX 1080 Ti GPU has been used.

*A. Fine-tuning*

Fine-tuning process has been exactly the same as training of SDANet which is based on 3 stage Cascade Mask R-CNN. The backbone network of the model is ResNet-101. Similar to pre-training, images' width is approximately resized to 500 while retaining the aspect ratio. The batch size is of size 2 and the optimizer is gradient descent with the initial training step of 0.02 and momentum of 0.9 and weight decay of 0.0001. A 12 gigabyte GeForce GTX 2080 Ti GPU has been used for fine-tuning process.

*B. Main Results*

Our first category of experimental results refers to the comparison of our proposed model performance with that of random initialization for different metrics and test cases. Based on our findings, given in TABLE I. , our model SegLoc has been able to outperform the random initialization in most of the experiments; More specifically, our model could outperform random initialization by 3 to 6 percent in almost all metrics in both tasks of object detection and semantic segmentation in test cases of Easy and Hard. This suggests the ability of SegLoc method in extraction of meaningful and beneficial features for tasks of object detection and semantic segmentation. Nevertheless, one of our findings raise some questions, that is, in the object detection task, our model underperformed in Hidden test case by 1 to 2 percent, while Hidden test case has been originally introduced to evaluate the generalization ability of models in detecting intentionally hidden forbidden objects (e.g., hiding forbidden objects by being wrapped by metal wires). Taking into consideration that we have been trying to further enhance the generalization ability of our model by decreasing lower limit of composition coefficient value range in order to reinforce our model to solve more challenging detections, as if pasted segments' objects were intentionally hidden under pile of other objects, the obtained outcome seems surprising specifically when considering the outperformance in other two test cases. Regardless, this underperformance may be attributed to the following two reasons, first being overfitting of our model on training dataset, as Easy and Hard test cases contain segments of the same quality as that of training dataset in contrast to Hidden test case and the second being some subtle challenges which we had encountered during experiments by which we were obliged to omit batch normalization layers for the pre-training model which were replaced again for downstream task model, which is further discussed in the discussion section.

Our second category of experimental results, given in TABLE I. , refers to the comparison of our model's performance with that of supervised initialization. Indeed, in the supervised initialization, the ResNet-101 backbone was taken from publicly available Pytorch version which was pre-trained for 100 epochs on ImageNet dataset, as was the case for baseline paper [17]. The evaluation process of this category of experiments are the same as previous category, that is, three test cases of Easy, Hard, and Hidden along with similar evaluation metrics were considered.

Based on our experiments, in object detection task, our proposed model underperformed in almost all cases compared to supervised initialization model; more specifically, our model underperformed in Easy and Hard test cases for about 9 to 17 percent while in hidden test case the underperformance was even more noticeable which again raises the same questions as that of previous experiment. Similarly, for semantic segmentation task, our model underperformed in Easy and Hard test cases for about 9 to 17 percent while in hidden test case the underperformance was even more noticeable which again raises the same questions as that of previous experiment. Similarly, for semantic segmentation task, our model underperformed the supervised initialization except that differences were up to 4 percent lower than that of object detection task. Nevertheless, the results obtained from supervised initialization were unexpected as the supervised initialization was pre-trained on a dataset consisting of images with absolutely different domains as opposed to that of our dataset.

TABLE I. OBJECT DETECTION AND SEMANTIC SEGMENTATION RESULTS ON PIDRAY DATASET TEST SET FOR OUR MODEL (SEGLOC), RANDOM INITIALIZATION, AND SUPERVISED INITIALIZATION

|  |  | Detection | | | Segmentation | | |
| --- | --- | --- | --- | --- | --- | --- | --- |
|  |  | *Easy* | *Hard* | *Hidden* | *Easy* | *Hard* | *Hidden* |
| Ours (SegLoc) | AP @ [ IoU = 0.5 : 0.95 \| maxDets = 100 ] | 0.537 | 0.528 | 0.238 | 0.470 | 0.432 | 0.192 |
|  | AP @ [ IoU = 0.50 \| maxDets = 100 ] | 0.698 | 0.736 | 0.371 | 0.684 | 0.695 | 0.341 |
|  | AP @ [ IoU = 0.75 \| maxDets = 100 ] | 0.611 | 0.599 | 0.256 | 0.551 | 0.476 | 0.201 |
|  | AR @ [ IoU = 0.5 : 0.95 \| maxDets = 1 ] | 0.631 | 0.558 | 0.380 | 0.546 | 0.468 | 0.304 |
|  | AR @ [ IoU = 0.5 : 0.95 \| maxDets = 10 ] | 0.650 | 0.626 | 0.422 | 0.560 | 0.515 | 0.332 |
|  | AR @ [ IoU = 0.5 : 0.95 \| maxDets = 100 ] | 0.650 | 0.626 | 0.422 | 0.560 | 0.515 | 0.332 |
| Random Initialization | AP @ [ IoU = 0.5 : 0.95 \| maxDets = 100 ] | 0.491 | 0.467 | 0.255 | 0.428 | 0.375 | 0.199 |
|  | AP @ [ IoU = 0.50 \| maxDets = 100 ] | 0.652 | 0.674 | 0.393 | 0.639 | 0.630 | 0.352 |
|  | AP @ [ IoU = 0.75 \| maxDets = 100 ] | 0.556 | 0.525 | 0.279 | 0.501 | 0.400 | 0.200 |
|  | AR @ [ IoU = 0.5 : 0.95 \| maxDets = 1 ] | 0.601 | 0.510 | 0.390 | 0.516 | 0.419 | 0.303 |
|  | AR @ [ IoU = 0.5 : 0.95 \| maxDets = 10 ] | 0.623 | 0.583 | 0.435 | 0.533 | 0.469 | 0.331 |
|  | AR @ [ IoU = 0.5 : 0.95 \| maxDets = 100 ] | 0.623 | 0.583 | 0.435 | 0.533 | 0.469 | 0.331 |
| Supervised Initialization | AP @ [ IoU = 0.5 : 0.95 \| maxDets = 100 ] | 0.709 | 0.641 | 0.482 | 0.595 | 0.520 | 0.361 |
|  | AP @ [ IoU = 0.50 \| maxDets = 100 ] | 0.826 | 0.814 | 0.632 | 0.818 | 0.778 | 0.592 |
|  | AP @ [ IoU = 0.75 \| maxDets = 100 ] | 0.782 | 0.716 | 0.542 | 0.707 | 0.590 | 0.408 |
|  | AR @ [ IoU = 0.5 : 0.95 \| maxDets = 1 ] | 0.773 | 0.636 | 0.565 | 0.653 | 0.528 | 0.433 |
|  | AR @ [ IoU = 0.5 : 0.95 \| maxDets = 10 ] | 0.789 | 0.707 | 0.591 | 0.663 | 0.575 | 0.450 |
|  | AR @ [ IoU = 0.5 : 0.95 \| maxDets = 100 ] | 0.789 | 0.707 | 0.591 | 0.663 | 0.575 | 0.450 |

Despite the fact that determining how much each factor may be responsible in these findings is beyond the scope of this work, however, some issues worth considering. First of all, the pre-trained model in supervised pre-training was obtained by performing an object detection task which contained all modules and connections existed in downstream model, more importantly, the supervised model for ImageNet has obtained by many experiments and finding the best set of hyperparameters by which the model was trained for 100 epochs on a curated dataset of ImageNet which contains 1 million of images. In contrast, our model was pre-trained for about 30 epochs on much smaller scale of dataset in contrastive learning manner.

Moreover, same as previous category of experiments, again we had to encounter the issue of discarding batch normalization layers of ResNet-101 existed in downstream model which further deteriorates the performance as there will be some layers with randomly initialized weights among pre-trained weights of other layers in ResNet-101. In the next section, we have discussed the problem with batch-normalization layer along with some findings which may be of crucial importance in the process of learning.

*C. Discussion*

Regarding the main challenges of this research, limitation of GPU memory was the most noticeable one whereby considering a larger batch size was not feasible, which is of high importance in efficacy of contrastive learning along with the size of dictionary. Additionally, as a result of this limitation, in order to practically increase the learning batch size, we resorted to gradient accumulation trick forcing us to discard batch normalization layer during pre-training since combination of batch normalization layers with gradient accumulation trick adversely affects training process.

Moreover, to reach the best configuration, we had conducted multiple ablation experiments to evaluate effect of different variables, namely number of frozen levels of backbone and dictionary initialization.

In case of frozen levels, our experiments have shown that the obtained results are superior when neither of the levels or layers are frozen. Also, more importantly, we have observed that dictionary initialization has played a crucial role in convergence of our pre-training model; indeed, we have observed that randomly initialized dictionary converged much slower or did not converge at all, in the horizon of 30 epochs in the pre-training phase compared to when the dictionary was initialized by keys obtained from initial state of the model (i.e., right before the onset of pre-training). In other words, in this case, the dictionary was initialized by the keys obtained by feeding the pre-training synthesized images to randomly initialized pre-train model. This may be rooted in the fact that it takes several steps for the model's dictionary to be filled with meaningful key vectors and get rid of initial random vectors.

VI. CONCLUSION

In this article, we have proposed a new scheme in concept of self-supervised learning to harness the unlabeled specialized images for dense prediction tasks. More specifically, we have integrated our model SegLoc into a specialized task with a domain different from typical natural domain employed in most self-supervised learning methods. We considered the task of prohibited items detection in x-ray images and conducted multiple experiments which have proven our model's capacity in capturing meaningful features from synthesized images, thereby decreasing the gap between random and supervised initialization of neural networks. Thanks to our scheme, we have addressed one of the major issues of false negative pairs in contrastive learning by applying an adaption into the dictionary maintenance of MoCo method. Also, we have discussed the crucial role of dictionary initialization in proper convergence during pre-training. Lastly, SegLoc's computational cost does not exceed that of InsLoc's, while more comprehensive evaluations in terms of quality and computational cost in comparison to other SOTA models are deferred to future studies.

REFERENCES

[1] J. Devlin, "Bert: Pre-training of deep bidirectional transformers for language understanding," arXiv preprint arXiv:1810.04805, 2018.

[2] T. Brown et al., "Language models are few-shot learners," Advances in neural information processing systems, vol. 33, pp. 1877-1901, 2020.

[3] C. Doersch, A. Gupta, and A. A. Efros, "Unsupervised visual representation learning by context prediction," in Proceedings of the IEEE international conference on computer vision, 2015, pp. 1422-1430.

[4] S. Gidaris, P. Singh, and N. Komodakis, "Unsupervised representation learning by predicting image rotations," arXiv preprint arXiv:1803.07728, 2018.

[5] M. Noroozi and P. Favaro, "Unsupervised learning of visual representations by solving jigsaw puzzles," in Computer Vision–ECCV 2016: 14th European Conference, Amsterdam, The Netherlands, October 11-14, 2016, Proceedings, Part VI, 2016: Springer, pp. 69-84.

[6] D. Pathak, P. Krahenbuhl, J. Donahue, T. Darrell, and A. A. Efros, "Context encoders: Feature learning by inpainting," in Proceedings of the IEEE conference on computer vision and pattern recognition, 2016, pp. 2536-2544.

[7] R. Santa Cruz, B. Fernando, A. Cherian, and S. Gould, "Visual permutation learning," IEEE transactions on pattern analysis and machine intelligence, vol. 41, no. 12, pp. 3100-3114, 2018.

[8] J. Xu et al., "Noisy-as-clean: Learning self-supervised denoising from corrupted image," IEEE Transactions on Image Processing, vol. 29, pp. 9316-9329, 2020.

[9] K. He, H. Fan, Y. Wu, S. Xie, and R. Girshick, "Momentum contrast for unsupervised visual representation learning," in Proceedings of the IEEE/CVF conference on computer vision and pattern recognition, 2020, pp. 9729-9738.

[10] M. Caron, P. Bojanowski, A. Joulin, and M. Douze, "Deep clustering for unsupervised learning of visual features," in Proceedings of the European conference on computer vision (ECCV), 2018, pp. 132-149.

[11] M. Caron, I. Misra, J. Mairal, P. Goyal, P. Bojanowski, and A. Joulin, "Unsupervised learning of visual features by contrasting cluster assignments," Advances in neural information processing systems, vol. 33, pp. 9912-9924, 2020.

[12] L. Jing and Y. Tian, "Self-supervised visual feature learning with deep neural networks: A survey," IEEE transactions on pattern analysis and machine intelligence, vol. 43, no. 11, pp. 4037-4058, 2020.

[13] A. Jaiswal, A. R. Babu, M. Z. Zadeh, D. Banerjee, and F. Makedon, "A survey on contrastive self-supervised learning," Technologies, vol. 9, no. 1, p. 2, 2020.

[14] C. Yang, Z. Wu, B. Zhou, and S. Lin, "Instance localization for self-supervised detection pretraining," in Proceedings of the IEEE/CVF Conference on Computer Vision and Pattern Recognition, 2021, pp. 3987-3996.

[15] X. Chen, H. Fan, R. Girshick, and K. He, "Improved baselines with momentum contrastive learning," arXiv preprint arXiv:2003.04297, 2020.

[16] J. Deng, W. Dong, R. Socher, L.-J. Li, K. Li, and L. Fei-Fei, "Imagenet: A large-scale hierarchical image database," in 2009 IEEE conference on computer vision and pattern recognition, 2009: Ieee, pp. 248-255.

[17] B. Wang, L. Zhang, L. Wen, X. Liu, and Y. Wu, "Towards real-world prohibited item detection: A large-scale x-ray benchmark," in Proceedings of the IEEE/CVF international conference on computer vision, 2021, pp. 5412-5421.

[18] C. Miao et al., "Sixray: A large-scale security inspection x-ray benchmark for prohibited item discovery in overlapping images," in Proceedings of the IEEE/CVF conference on computer vision and pattern recognition, 2019, pp. 2119-2128.

[19] J. Geiping et al., "A Cookbook of Self-Supervised Learning," arXiv preprint arXiv:2304.12210, 2023.

[20] A. Dosovitskiy et al., "An image is worth 16x16 words: Transformers for image recognition at scale," arXiv preprint arXiv:2010.11929, 2020.

[21] P. Bachman, R. D. Hjelm, and W. Buchwalter, "Learning representations by maximizing mutual information across views," Advances in neural information processing systems, vol. 32, 2019.

[22] T. Chen, S. Kornblith, M. Norouzi, and G. Hinton, "A simple framework for contrastive learning of visual representations," in International conference on machine learning, 2020: PMLR, pp. 1597-1607.

[23] J.-B. Grill et al., "Bootstrap your own latent-a new approach to self-supervised learning," Advances in neural information processing systems, vol. 33, pp. 21271-21284, 2020.

[24] M. Caron et al., "Emerging properties in self-supervised vision transformers," in Proceedings of the IEEE/CVF international conference on computer vision, 2021, pp. 9650-9660.

[25] A. Bardes, J. Ponce, and Y. LeCun, "Variance-invariance-covariance regularization for self-supervised learning," ICLR, Vicreg, vol. 1, no. 2, 2022.

[26] E. Xie et al., "Detco: Unsupervised contrastive learning for object detection," in Proceedings of the IEEE/CVF International Conference on Computer Vision, 2021, pp. 8392-8401.

[27] Z. Dai, B. Cai, Y. Lin, and J. Chen, "Up-detr: Unsupervised pre-training for object detection with transformers," in Proceedings of the IEEE/CVF conference on computer vision and pattern recognition, 2021, pp. 1601-1610.

[28] F. Wei, Y. Gao, Z. Wu, H. Hu, and S. Lin, "Aligning pretraining for detection via object-level contrastive learning," Advances in Neural Information Processing Systems, vol. 34, pp. 22682-22694, 2021.

[29] C. Li et al., "Efficient self-supervised vision transformers for representation learning," arXiv preprint arXiv:2106.09785, 2021.

[30] F. Ghesu et al., "Self-supervised Learning from 100 Million Medical Images. arXiv 2022," arXiv preprint arXiv:2201.01283.

[31] S. Azizi et al., "Big self-supervised models advance medical image classification," in Proceedings of the IEEE/CVF international conference on computer vision, 2021, pp. 3478-3488.

[32] M. Vo, E. Yumer, K. Sunkavalli, S. Hadap, Y. Sheikh, and S. G. Narasimhan, "Self-supervised multi-view person association and its applications," IEEE transactions on pattern analysis and machine intelligence, vol. 43, no. 8, pp. 2794-2808, 2020.

[33] S. Shurrab and R. Duwairi, "Self-supervised learning methods and applications in medical imaging analysis: A survey," PeerJ Computer Science, vol. 8, p. e1045, 2022.

[34] K. He, G. Gkioxari, P. Dollár, and R. Girshick, "Mask r-cnn," in Proceedings of the IEEE international conference on computer vision, 2017, pp. 2961-2969.

[35] Z. Cai and N. Vasconcelos, "Cascade r-cnn: Delving into high quality object detection," in Proceedings of the IEEE conference on computer vision and pattern recognition, 2018, pp. 6154-6162.